\documentclass[letterpaper, 10 pt, conference, onecolumn]{IEEEtran}
\IEEEoverridecommandlockouts

\usepackage{cite}
\usepackage{amsmath,amssymb,amsfonts}
\usepackage{algorithmic}
\usepackage{graphicx}
\usepackage{textcomp}
\usepackage{xcolor}
\usepackage{float}
\usepackage{url} 
\def\BibTeX{{\rm B\kern-.05em{\sc i\kern-.025em b}\kern-.08em
    T\kern-.1667em\lower.7ex\hbox{E}\kern-.125emX}}
\begin{document}

\title{A Novel Graph-Sequence Learning Model for Inductive Text Classification}
\author{
\IEEEauthorblockN{Zuo Wang\textsuperscript{1*} and Ye Yuan\textsuperscript{1}}
\IEEEauthorblockA{\textsuperscript{1}{\textit{College of Computer and Information Science, Southwest University, Chongqing, China}}\\ 
wz20011013@email.swu.edu.cn\textsuperscript{*}, yuanyekl@swu.edu.cn
}
}



\maketitle

\begin{abstract}
Text classification plays an important role in various downstream text-related tasks, such as sentiment analysis, fake news detection, and public opinion analysis.
Recently, text classification based on Graph Neural Networks (GNNs) has made significant progress due to their strong capabilities of structural relationship learning.
However, these approaches still face two major limitations.
First, these approaches fail to fully consider the diverse structural information across word pairs, e.g., co-occurrence, syntax, and semantics.
Furthermore,  they neglect sequence information in the text graph structure
information learning module and can not classify texts with new words and relations.
In this paper, we propose a Novel Graph-Sequence Learning Model for Inductive Text Classification (TextGSL) to address the previously mentioned issues.
More specifically, we construct a single text-level graph for all words in each text and establish different edge types based on the diverse relationships between word pairs. Building upon this, we design an adaptive multi-edge message-passing paradigm to aggregate diverse structural information between word pairs.
Additionally, sequential information among text data can be captured by the proposed TextGSL through the incorporation of Transformer layers. 
Therefore, TextGSL can learn more discriminative text representations. 
TextGSL has been comprehensively compared with several strong baselines.
The experimental results on diverse benchmarking datasets demonstrate that TextGSL outperforms these baselines in terms of accuracy.
\end{abstract}

\section{Introduction}

Text classification\cite{minaee2021deep} has achieved extensive applications across various text-related downstream tasks, such as sentiment analysis\cite{wankhade2022survey}, spam filtering \cite{ manita2023efficient}, topic detection in social media \cite{panchendrarajan2023topic}, and intelligent question-answering systems\cite{yang2024application,tang2021intelligent}. 
With the rapid development of Graph Neural Networks (GNNs)\cite{wu2020comprehensive,kipf2016semi, velivckovic2017graph, HE2024104129,10819283}, graph-based text classification approaches have achieved outstanding results.
Existing graph-based approaches for text classification follow a two-step strategy.
First, they construct different graph structures (eg., corpus-level, text-level, sentence-level graphs) for the text data.
Then, they apply Graph Neural Networks (GNNs)\cite{10027700,9885025,9159907} to capture the complex structural information in the text data.
In contrast to traditional methods, graph-based approaches learn text representations from the rich structures of non-Euclidean spaces, which are converted from the text data in the form of simple sequence structures.
For instance, 
TextGCN\cite{yao2019graph} constructs a corpus-level heterogeneous graph and uses graph convolutional networks (GCN) to learn complex structural information between texts.  
TensorGCN \cite{liu2020tensor} employs higher-order tensors to represent text graph data, leveraging both intra-graph and inter-graph propagation to integrate richer contextual information.
Recently, pre-trained models have also been widely applied in text classification tasks.
These approaches leverage pre-trained models to obtain text embeddings, which are then used as node features in the graph neural networks.
BertGCN\cite{lin-etal-2021-bertgcn} learns text representations by combining the features of pre-trained BERT\cite{devlin2019bert} and graph convolutional neural network.
BertGACN\cite{10105710} integrates the BERT model with a dual-tower graph neural network model, further improving text classification performance.

However, these approaches fail to fully consider the diverse structural dependencies between word pairs. 
Moreover, they convert sequential text data into non-Euclidean graph structures, but neglect sequence information in the text graph structure information learning module.
For instance, TextSSL\cite{piao2022sparse} learns sparse structural information of graphs by dynamically constructing edges between sentences, but this method fails to capture diverse information between word pairs. 
TextING\cite{zhang-etal-2020-every} constructs a text-level graph based on a sliding window, which captures only co-occurrence relationships between word pairs. 
TensorGCN \cite{liu2020tensor} considers different relationships at the corpus-level graph, but constructing the tensor graph has high complexity. These graph-based\cite{he2021multi,8766847,CHEN2025103297,he2022neighborsworthattendingto}
approaches all neglect sequential information in the graph structure learning module.
Although pre-trained \cite{wang2023large} models can enhance the learning of long-range sequential information, they require more computational resources.

To address the above limitations, we propose a Graph-Sequence Learning Model for inductive Text classification (TextGSL), which can effectively leverage diverse structural information and sequential information to learn more discriminative text representations. 
Specifically, for each document, we construct a single text-level graph containing all words based on the different relationships (e.g., co-occurrence, syntax, semantics).
These relationships are incorporated as edge features, enabling the model to learn the diverse structural information from the text-level graph. 
To capture the diverse structural information in text-level graphs, we designed an adaptive multi-edge message-passing paradigm. 
In this new paradigm, we first compute edge weights for aggregating neighborhood information based on similarity and dissimilarity between word nodes.
Second, for different relationships, we use adaptive parameters as edge weight scaling factors to dynamically measure the importance of different relationships and obtain the diverse neighborhood structural information for the central word node.
Finally, Gated Graph Neural Network
(GG-NN) \cite{li2016gated} is applied to update the hidden states at each layer. 
To address the over-smoothing problem that fundamentally limits the capacity of graph neural networks, the proposed TextGSL additionally adopts Transformer\cite{vaswani2017attention} layers [20] to capture the long-range, and sequential information hidden in the text data.
Then, we employ a Bi-GRU\cite{pan2020chinese} module to fuse the diverse structural and sequential features learned by the GG-NN with the proposed multi-edge message-passing paradigm and transformer layer.
Finally, an attention mechanism is applied to learn graph-level representations.
Based on the fusion of diverse structural information and long-range sequential information, our model can learn more discriminative text representations. 

The main contributions of this paper are as follows:


\begin{itemize}
    \item We construct a text-level graph based on the different relationships (e.g., co-occurrence, syntax, semantics) between word pairs, and design an adaptive multi-edge messaging-paradigm.
    This messaging-paradigm can dynamically measure the importance of different relationships between word pairs, thus enabling the integration of diverse structural information at the text-level graph.
 
    \item We propose a Graph-Sequence learning model for Inductive Text Classification (TextGSL). Our model can learn more discriminative text representations by integrating diverse structural information and long-range sequential
    information. 

    \item We conduct a series of experiments that compare the proposed TextGSL with other strong baselines on text classification. The experimental results demonstrate that TextGSL can outperform all
    compared baselines on all test datasets.
 
\end{itemize}

\section{Related works}
In this section, we briefly review existing approaches to text classification that can be categorized into three classes, i.e., classical machine learning-based methods, deep learning-based methods, and graph neural network-based\cite{11114958} methods.

\subsection{Classical Machine Learning-Based Methods}
In empirical machine learning-based text classification approaches, the process for text classification typically follows two steps: first, they manually extract features from the text, and then use shallow classification models for text prediction, such as Support Vector Machines(SVM)\cite{mammone2009support}, KNN\cite{tan2006effective}, and Fast Decision Tree (FDT)\cite{10.1145/234313.234346}.
Common traditional feature extraction methods include bag of words (BOW)\cite{zhang2010understanding}, Term Frequency-Inverse Document Frequency (TF-IDF)\cite{aizawa2003information}, and N-gram model.
However, these traditional feature extraction methods are inefficient and cannot learn deep contextual semantic information of the text.

\subsection{Deeping Learning-Based Methods}
Text classification approaches based on deep learning models have a more complex structure, but they do not require manual feature extraction and can learn shallow semantic information.
Common deep learning\cite{9357419} based text classification approaches include convolutional neural networks (CNN)\cite{kalchbrenner-etal-2014-convolutional}, recurrent neural networks (RNN)\cite{cho-etal-2014-learning}, long short-term memory networks (LSTM), and related attention mechanisms models\cite{dhingra2017gated}. 
Although these approaches have made progress in text classification tasks, they are difficult to understand deeper semantic information and complex structural relationships in text data.

\subsection{Graph Neural Network-Based Methods}
The success of graph neural networks\cite{zhou2024differentiable,wu2019simplifying,10027700,10035508,9764654} in learning non-Euclidean structured data has provided significant inspiration for advancing text classification approaches\cite{minaee2021deep}. 
Graph neural networks\cite{10265117,10830921} aggregate neighbor information through the message-passing paradigm \cite{he2021learning}, i,e, 
graph convolutional neural network \cite{kipf2016semi} makes use of the Graph Laplacian derived from the adjacency matrix to aggregate neighboring information for representation learning. 
Gated Graph Neural Network (GG-NN)\cite{li2016gated} incorporates a gated mechanism into the message-passing process to address the limitations of earlier GNNs\cite{11151980} in modeling long-range dependencies and complex state transitions. 
More and more researchers are using graph neural networks to learn complex relationships in text data.
Some studies build a corpus-level graph to learn the global information. For instance, 
InductGCN\cite{Wang2022InducTGCNIG} proposed the first inductive text classification framework based on corpus-level graph learning.
HyperGAT \cite{ding-etal-2020-less} employs attention mechanisms\cite{10.1145/3719295} to dynamically focus on important nodes and hyperedges within the textual hypergraph.
CGA2TC\cite{yang2022contrastive} designs a text classifier based on two contrastive perspectives, but this design increases computational overhead. 
Graph Fusion Network (GFN)\cite{DAI2022107659} integrates multiple different corpus-level graph feature representations to obtain the final text representation.
HEGAT\cite{linmei-etal-2019-heterogeneous} learns the relationships between nodes in a corpus-level graph through Heterogeneous Graph Attention Network (HAN).
Although these corpus-level graph-based text classification approaches can effectively capture global information from text data, they cannot perform inductive learning on new texts, and building a corpus-level graph would increase computational complexity for long-text data. 

Instead, other studies have constructed a text-level graph containing all words for each document. 
KGAT \cite{10.1007/978-3-031-17120-8_51} constructed text-level graphs and then uses an enhanced multi-head
graph attention network to capture semantic and structural information.
TextLevelGNN\cite{huang-etal-2019-text} reduces memory consumption through globally shared node features and edge weights in graph structural information learning.
Recently, pre-trained models \cite{qiu2020pre} for text classification have also achieved significant progress in text classification tasks. 
AGGNN \cite{deng2022text} uses an attention-gated recurrent network to aggregate word node information, which can effectively capture the semantic relationships between words in text-level graphs. 

\begin{figure}[t]\renewcommand{\arraystretch}{1.5}
	\centering
\includegraphics[width=0.60\textwidth]{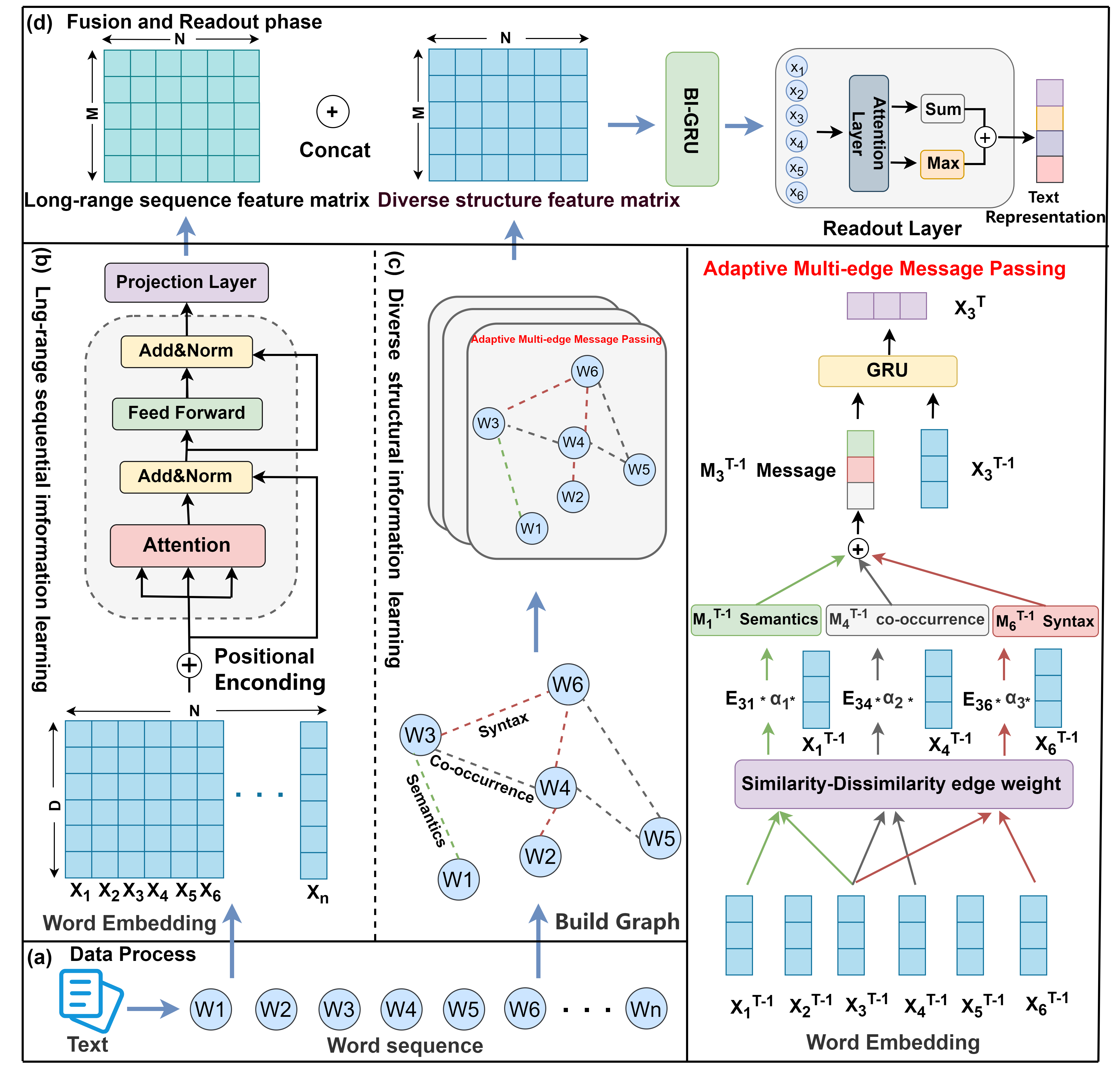}
	\caption{The overall framework of the proposed TextGSL for inductive text classification. \textbf{(a)} is the data processing phase, which converts text into a sequence of words. \textbf{(b)} is the long-range sequential information learning module. \textbf{(c)} is the diverse structural information learning module.  \textbf{(d)} integrates diverse structural information with local sequence information, and all nodes aggregate to the ultimate document graph
    representation with an attention mechanism in the readout
    phase.}
	\label{frame}
\end{figure}

\section{Formulation OF TEXTGSL}
In this section, we propose a Graph-Sequence learning model for Inductive Text Classification (TextGSL). 
TextGSL includes three parts: the long-range sequential information learning module, the diverse structural information learning module, fusion and readout phase.
In the local sequence information learning module, we use a transformer as the encoder to effectively capture
long-range sequential dependencies in text data.
In the diverse structural information learning module, we construct a text-level graph based on the different relationships (e.g., co-occurrence, syntax, semantics) between word pairs, and design an adaptive multi-edge messaging-paradigm to dynamically aggregate neighbor information from different edge relationships.
In the fusion and readout phase, we use Bi-GRU to integrate long-range sequential information with diverse structural information.
Then, all nodes are aggregated into the ultimate document graph representation with an attention mechanism.
Fig.\ref{frame} shows the overall architecture of this model. 
The following text will elaborate on the specific implementation of nations, long-range sequential information learning, diverse structural information learning, fusion and readout phase.

\subsection{Nations}
Before introducing the proposed TextGSL, we list the necessary notations and preliminaries used in this paper.
We use $G = \left \{E, V, \mathbf X \right \}$ to represent a text-level graph. 
$V \in \left \{ v_{1},v_{2}...v_{n} \right \}$ denotes the word node set.
$E\in \left \{  E_{co-occur} ,E_{syntax} ,E_{semantics}\right \}$ denotes the relationships between word pairs.
$\mathbf X\in N\times D = \left \{ x_{1},x_{2}...x_{n} \right \}$ denotes the input feature matrix of the word, where $x_{n}$ denotes the feature for word node $v_{n}$ and $D$ denotes the dimension of feature.
$Y_{i}$ denotes the class label to which the document belongs. We use $ N_{i} $ to denote the set of neighboring nodes of node $i$. $\mathbf A\in \left \{ 0,1 \right \}^{n\times n}$ represents the adjacency matrix of $G$. 
$\mathbf W$ represents the matrices of learnable weights.

\subsection{Long-range Sequential Information Learning}
Existing graph-based text classification approaches neglect sequential information in the graph structure learning module.
In addition, the over-smoothing problem fundamentally limits the capacity of graph neural networks to capture long-range dependencies.
To address these issues, we use a transformer \cite{vaswani2017attention} as the encoder for long-range, and sequential information.
Long-range sequential information and diverse structural information are combined to enhance text classification performance.
We obtain the long-range sequence feature matrix through the transformer encoder:
\begin{equation}\label{eq:pe}
\begin{aligned}
    &\mathbf{X} = \mathbf{X} + \text{PE}(\mathbf{X}), \\[4pt]
    &\text{PE}(\text{pos}, 2i) = \sin \left(\text{pos} /1000^{2i/d_{\text{model}}}\right), \\[4pt] 
    &\text{PE}(\text{pos}, 2i+1) = \cos\left(\text{pos} 1000^{2i/d_{\text{model}}}\right), \\[4pt]
    &\mathbf H^{''} =  \text{Encoder}(\mathbf X), \\[4pt]
    &\mathbf{X}_{seq} = f(\mathbf{H^{''}}),\\
\end{aligned}
\end{equation}

where $\text{PE}$ denotes position encoder.  $d_{\text{model}}$ are the embedding dimension of a word.
$\mathbf X_{} \in \mathbb{R}^{N \times D}$ denotes word feature matrix. $\mathbf X_{seq} \in \mathbb{R}^{N \times M}$ denotes long-range sequence feature matrix through encoder. $f$ denotes a projection Layer.

The specific learning process of the long-range sequential information encoder is as follows:  

\begin{equation}\label{eq:projection}
\begin{aligned}
    & \mathbf{Q} = \mathbf{X}\mathbf{W}_Q,  
    \mathbf{K} = \mathbf{X}\mathbf{W}_K,  
    \mathbf{V} = \mathbf{X}\mathbf{W}_V,\\[4pt]
    & \mathbf{Z} = \text{Attention}(\mathbf Q, \mathbf K, \mathbf V)=\text{Softmax}\left(\frac{\mathbf{Q}\mathbf{K}^{T}}{\sqrt{d_k}}\right) \mathbf{V},\\[4pt]
    &\mathbf{H}_{\text{}} = \text{LayerNorm}\left(\mathbf{X} + \text{Dropout}(\mathbf{Z}\right)),\\[4pt]
    &\mathbf{H^{'}} = \text{ReLU} \left(\mathbf{H}_{\text{}} \mathbf{W}_1 + \mathbf{b}_1\right)\mathbf{W}_2 + \mathbf{b}_2,\\[4pt]
    &\mathbf{H^{''}}_{\text{}} = \text{LayerNorm}(\mathbf{H} + \text{Dropout}(\mathbf{H^{'}})),\\
\end{aligned}
\end{equation}





where $\mathbf W_Q,\mathbf W_K, \mathbf W_V, \mathbf W_1,\mathbf W_2$ are learnable parameter matrices. $b_1, b_2$ are bias. $ \mathbf{F_{seq}}\in R^{N \times M } $ denotes is the local sequence feature matrix output by the encoder.

\subsection{Diverse Structural Information Learning}
\begin{figure}[t]\renewcommand{\arraystretch}{1.5}
	\centering
\includegraphics[width=0.45\textwidth]{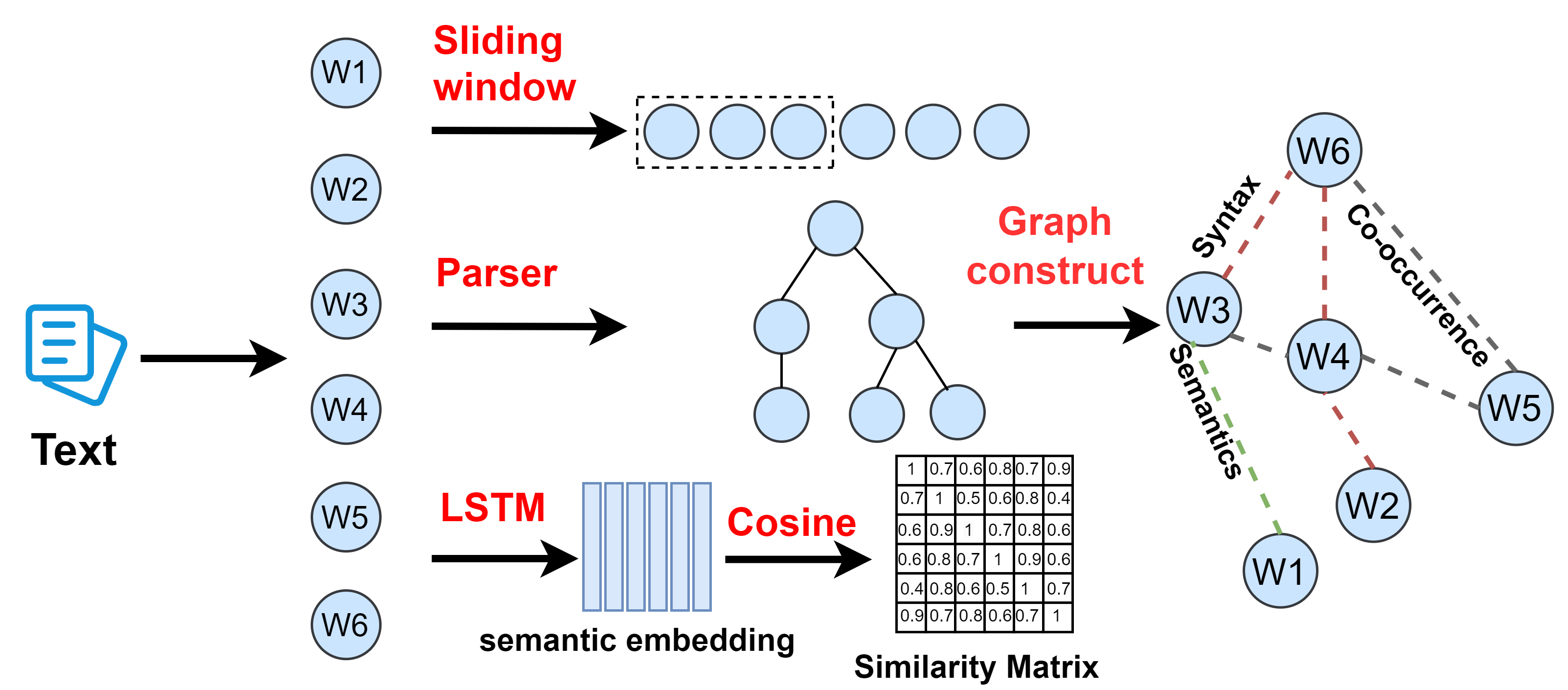}
	\caption{Text-level graph construction based on co-occurrence, syntax, and semantics relationships.}
	\label{graph}
\end{figure}

Most graph-based approaches for text classification do not fully consider the diverse structural information between word pairs in their graph structural information learning modules. 
Most of them constructed text-level graphs based on co-occurrence relationships, without considering the diverse relationships between word pairs.
Therefore, we consider the relationships between different word pairs into a single text-level graph, extracting co-occurrence, syntax, and semantic relationships as edge features.
The co-occurrence relationship describes the associations between word pairs that appear within a sliding window.
According to previous studies\cite{zhang-etal-2020-every}, in graph-based text classification approaches, sliding windows are widely used in text-level graph construction. 
For syntax relationships, we use Stanza1 as a parser to extract syntactic dependency relationships between word pairs, such as subject-verb relationships.
For semantic relationships, we refer to TensorGCN \cite{liu2020tensor} and use a pre-trained LSTM to obtain contextual embeddings for each word. Next, we calculate the cosine similarity between word pairs to construct semantic relationships. 
Fig.\ref{graph} shows the process of constructing the text-level graphs based on co-occurrence,
syntax and semantics relationships.

Besides, we have designed an adaptive multi-edge message-passing paradigm to dynamically measure the importance of different relationships in the graph structural information learning module. 
More specifically, we jointly calculate the edge weights of aggregated neighbor information based on similarity and dissimilarity. 
Then, for each relationship, an adaptive parameter is set as the edge weight scaling factor to ultimately obtain the neighbor message.
Finally, a node can receive information from its neighbor nodes and then merge with its last time step representation to update each node’s hidden state.
This paradigm can dynamically learn the importance of relationships between word pairs. 
The edge weight between node word $i$ and node word $j$ is calculated as follows:
\begin{equation}\label{edeg}
\begin{aligned}
&\mathbf S_{ij}^{t-1}= \mathbf A_{ij}\cdot [\mathbf \alpha (\mathbf X_{i:}^{t-1} || \mathbf X_{j:}^{t-1})^{T}],\\[4pt]
&\mathbf D_{ij}^{t-1}= \mathbf A_{ij}\cdot \mathbf |\mathbf X_{i}^{t-1} - \mathbf X_{j}^{t-1}|^2,\\[4pt]
&\mathbf E_{ij}^{t-1}  = e^{(\mathbf S_{ij}^{t-1}-\beta \mathbf D_{ij}^{t-1})},\\
\end{aligned}
\end{equation}
where $\alpha$ is a vector of learnable parameters. $\beta$ is a learnable parameter to balance the relative importance of similarity and dissimilarity. $\mathbf E_{ij}^{t-1} $ denotes edge weight between word $i$ and word $j$ at time step $t-1$. 

According to the Eq.\ref{edeg}, we obtain the edge weight matrix $\mathbf E$. 
Then, in the graph structural information learning module, we set adaptive parameters for each edge type to dynamically measure the importance of different relationships. 
Finally, the neighbor information and the last time step $\mathbf X^{t-1}$ representation are fed to GRU \cite{li2016gated} to update the hidden state of $\mathbf X^{t}$.
The specific process of the propagation model is as follows:
\begin{equation} \label{14}
\begin{aligned}
&\mathbf a_i^t = \sigma (\sum_{j \in \mathcal{N}(i)} \gamma_{\ell} \cdot \mathbf{E}_{ij}^{t-1} \mathbf{X}_j^{t-1} ), \quad \ell \in \{ \text{co}, \text{syn}, \text{sem}\},\\[4pt]
&\mathbf z^t = \sigma(\mathbf W_z \mathbf a^t + \mathbf U_z \mathbf X^{t-1}) + \mathbf b_z ,\\[4pt]
&\mathbf r^t = \sigma(\mathbf W_r \mathbf a^t + \mathbf U_r \mathbf X^{t-1}) + \mathbf b_r,\\[4pt]
&\tilde{\mathbf X}^t = \tanh(\mathbf W_h \mathbf a^t + \mathbf U_h (\mathbf r^t \odot \mathbf X^{t-1}) + \mathbf b_h),\\[4pt]
&\mathbf X_{}^t = (1 - \mathbf z^t) \odot \mathbf X^{t-1} + \mathbf z^t \odot  \tilde{\mathbf X}^t,\\
\end{aligned}
\end{equation}



where $\mathbf a_i^t $ denotes neighbor messages generated based on different relationships at time step t. 
$\sigma $ denotes activation function. $\gamma_{\ell} $ denotes adaptive parameters for co-occurrence, syntax, and semantic relationships.
$z$ and $r$ are the update
and the reset gates. $z$ decides whether to update the hidden state. $r$ decides how much information from the $t-1$ time step to retain. $\mathbf W,\mathbf U$ are learnable parameters matrices. $\mathbf b_z,\mathbf b_r,\mathbf b_h$ are bias. $\odot$ denotes element-wise (Hadamard) product operator. $\mathbf X^t$ are obtained from the previous and the reserved hidden states, using the coefficients returned by the update gate.

\subsection{Fusion and Readout Phase}
This module aims to integrate long-range sequential information with diverse structural information to further obtain high-quality word node embeddings. 
we obtained two matrices $\mathbf X_{seq},\mathbf X_{str}$ in the long-range sequential information learning module and the diverse structural information learning module. 
Then, we use BI-GRU to integrate Long-range sequential information with diverse structural information. BI-GRU can simultaneously process inputs from both forward and backward directions. 
we use the following strategy to fuse long-range sequence feature  matrix and diverse structure feature matrix:
\begin{equation}
\mathbf X_{out} = \text{BI-GRU}(\mathbf X_{seq}  \oplus  \mathbf X_{str} )
\end{equation}
where $\mathbf X_{seq} \in \mathbb{R}^{N \times M}, \mathbf X_{str} \in \mathbb{R}^{N \times M}$ are the long-range sequence feature matrix and the diverse structure feature matrix, respectively.
$\mathbf X_{out} \in \mathbb{R}^{N \times M}$ denotes the final representation matrix of each word nodes.

In the readout phase, node representations will be aggregated into graph-level representations, which will then be used to predict text labels. 
We use the attention mechanisms, max-pooling, and mean-pooling to obtain the final text representation. The detailed process is as follows:
\begin{equation} \label{attention}
\begin{aligned}
    &\alpha_v = \text{attention}( f_{1}(\mathbf X^{}_{out})),
\\
&\mathbf X_v = \alpha_v \odot \tanh(  f_{2} (\mathbf X^{}_{out})),\\
&\mathbf X_g = \text{MaxPooling}(X_1, \dots, X_v) + \frac{1}{|V|} \sum_{v \in V} X_v,\\
\end{aligned}
\end{equation}

where $f_{1}, f_{2}$  are two multilayer perceptrons (MLP). Eq. \eqref{attention} measures the importance of different words through a soft attention mechanism. $\text{tanh()} $ is a non-linear feature transformation.

We measure the impact of different word embeddings on the final text representation by calculating the attention between word nodes. Attention mechanism can effectively focus on words that are consistent with text labels.
Then, the labels are predicted by inputting the final text representation into the \text{softmax} function. 
We choose the cross-entropy function as the loss function:
\begin{equation}
\begin{aligned}
&\hat{Y} = \text{softmax}(\mathbf W \mathbf X_g + \mathbf b),\\
&\mathcal{L} = \sum_{c=1}^{C} y_{c} \log(\hat{Y}_{c}),
\end{aligned}
\end{equation}
where $\mathbf W$ denotes a learnable parameter matrix. $\mathbf b$ is bias. $\hat{Y}_{c}$ is a predictive label of the TextGSL model. $y_{c}$ is the ground truth one-hot encoding.

\section{EXPERIMENTS}
In this section, we verify the effectiveness of the proposed TextGSL with several strong baselines against the text classification task.
In addition, we conducted diverse relationship analyses in Section \ref{relation}.
Then, we conducted a series of ablation studies to investigate the key properties of the proposed local sequential information and diverse structural information learning module in Section \ref{ablation}.
Finally, we conducted learning ability stability analyses in Section \ref{train}.

\begin{table}[t]
\renewcommand{\arraystretch}{1.2}
\centering
\caption{Statistics of the test datasets.}
\label{dataset}
\normalsize
\begin{tabular}{c| c c c c c}
    \hline
    Dataset & R8 & R52 & MR & Ohsumed & 20NG \\
    \hline
    \hline
      \# Docs &7674 &9100 &10662 &7400 &18846 \\
      \# Train &5485 &6532 &7108 &3357 &11314\\  
      \# Test &2189 &2568 &3554 &4034 &7532\\
      \# Classes &8 &52 &2 &23 &20\\
      \# Vocab &7688 &8892 &18764 &14157 &42757\\
      Avg.Len &65.72 &69.82 &20.39 &135.82 &221.26\\
    \hline
\end{tabular}
\end{table}

\renewcommand{\arraystretch}{1.3}
\begin{table*}
\caption{Performance comparisons evaluated by ACC.\\
The best result is highlighted in bold, and the second-best result is underlined.}
\label{syjg}
\begin{center}
\normalsize \centering \begin{tabular}{p{2.3cm}|p{2.3cm} p{2.3cm} p{2.3cm} p{2.3cm} p{2.3cm}}
\hline
\textbf{Method}& \textbf{R8} & \textbf{R52} & \textbf{MR} & \textbf{Ohsumed} & \textbf{20NG}\\
\hline
\hline
TextING & 0.9814 \small ± 0.0021	& 0.9541 \small ± 0.0013	& 0.7863 \small± 0.0020 & 0.7044 \small± 0.0035	 & - \\

TensorGCN &	0.9804 \small ± 0.0008 & 0.9505 \small ± 0.0011 &	0.7791 \small ± 0.0007&	0.7011 \small ±   0.0024	&0.8774 \small ± 0.0005	\\

HyperGAT & 0.9797 \small ±  0.0023 & 0.9498 \small ±  0.0027 & 0.7832 \small ±  0.0027 & 0.6990 \small ±  0.0034 & 0.8662 \small ± 0.0016\\

DHTG & 0.9733 \small ± 0.0006 & 0.9393 \small ± 0.0010 & 0.7721 \small ± 0.0011 & 0.6880 \small ± 0.0033 & 0.8713 \small ± 0.0007	\\			

TextSSL & 0.9781 \small ± 0.0014 & 0.9548 \small ±  0.0026 & \underline{0.7974}\small  ± 0.0019 & 0.7059 \small ± 0.0038 & 0.8526 \small ± 0.0028\\	

TextGCN & 0.9707 \small± 0.0010 & 0.9356 \small ± 0.0018 & 0.7674 \small ± 0.0020 & 0.6836 \small ± 0.0056 & 0.8634 \small± 0.0009\\

CGA2TC & 0.9776 \small ± 0.0019 & 0.9447 \small ± 0.0016 & 0.7780 \small ± 0.0029 & 0.7062 \small ± 0.0045			 & -\\	

LDGCN & \underline{0.9832} \small ± 0.0005 & 	\underline{0.9571} \small± 0.0014 & 0.7825 \small ±  0.0011 & \underline{0.7085} \small±  0.0018 & \underline{0.8779} \small ± 0.0003\\				

GTG & 0.9722 \small ± 0.0010	 & 0.9446 \small ± 0.0008 & 0.7724  \small ± 0.0032 & 0.6972 \small ± 0.0011 & 0.8696 \small ± 0.0009\\	

\hline
TextGSL & \textbf{0.9834} \small ± 0.0006 & \textbf{0.9626} \small ± 0.0015 & \textbf{0.8127} \small ± 0.0021	 & \textbf{0.7216} \small ± 0.0030 & \textbf{0.8832} \small ± 0.0006 \\

\hline
\end{tabular}
\end{center}
\end{table*}

\subsection{Experimental Setup}
\paragraph{Datasets}
For a fair and comprehensive evaluation, we selected five real-world text datasets for our experiments, including R8, R52\footnote{\url{https://www.cs.umb.edu/~smimarog/textmining/datasets/.}}, MR\footnote{\url{http://www.cs.cornell.edu/people/pabo/movie-review-data/.}}, Ohsumed\footnote{\url{http://disi.unitn.it/moschitti/corpora.htm.}}, 20NG\footnote{\url{http://qwone.com/~jason/20Newsgroups/.}}.
R8 and R52 datasets come from the Reuters agency ,which contains financial news. R8 contains news texts from 8 different categories, while R52 contains news texts from 52 different categories.
MR is a movie review dataset used for binary sentiment classification, where each review consists of only one sentence.
The Ohsumed dataset comes from the medical literature database MEDLINE. Each text is an abstract of a medical study and can be classified into one of 23 disease categories.
The 20NG dataset consists of newsgroup documents, with each news item classified into 20 categories, covering a wide range of topics such as sports, politics, technology, and religion. We have summarized the characteristics of five datasets in Table~\ref{dataset}.
Following previous studies \cite{zhang-etal-2020-every}, we first eliminated stop words 
using the NLTK5 library.
Then, we preprocessed all datasets by removing non-English characters and low-frequency words that appeared less than five times. 
For the Mr dataset, stop words and low-frequency words were not removed due to the short length of the texts.

\paragraph{Baselines}
We compare our model with several advanced graph-based text classification baselines to validate the effectiveness of TextGSL.
TextING\cite{zhang-etal-2020-every} is a graph-based text classification method that constructs an independent text-level graph for each text and introduces GRU to learn embeddings of word nodes. It enables inductive learning of new words.
DHTG \cite{pmlr-v108-wang20l} proposes a novel trainable hierarchical topic graph, which is capable of further capturing semantic hierarchical variations from fine-grained to coarse-grained levels.
TensorGCN \cite{liu2020tensor} constructs a textual graph tensor and leverages both intra-graph and inter-graph propagation learning to integrate richer contextual information.
HyperGAT \cite{ding-etal-2020-less} constructs a hypergraph and then uses a graph attention network to learn the graph structure information.
TextSSL \cite{piao2022sparse} is a sparse structure learning method based on graph neural networks (GNNs) that optimizes graph structures through dynamic context dependencies.
TextGCN\cite{yao2019graph}  represents text data as a corpus-level graph and then uses a Graph Convolutional Neural Network to learn text representations.
CGA2TC \cite{yang2022contrastive} uses contrast learning to optimize the graph structure, further reducing redundant edges in the corpus-level graph.
LDGCN \cite{10.1007/s00530-023-01112-y} proposes a Local Discriminative Graph Convolutional Network to enhance the performance of text classification.
GTC \cite{liu2023transformer} incorporates part-of-speech (POS) information into the corpus graph, constructs edges between word nodes based on POS. In the layer-to-layer of GCN, the Transformer is used to extract the contextual and sequential information of the text. For fair comparisons, we configure all baselines based on their recommended settings. 

\paragraph{Experiment Settings}
For all test datasets, we randomly select 10\% from the training set as the validation set. We run the experiments on an NVIDIA GeForce RTX 4090 GPU.
In the TextGSL model, we set the node embedding size as 300, the hidden size of the transformer and GNN is 96, and initialize word embedding with the pre-trained Glove-300 \footnote{\url{http://nlp.stanford.edu/data/glove.6B.zip}}. For each experiment, we run 10 times and report the mean results. Dropout is set at 0.5 and 0.65 in the diverse structural information learning module and the long-range sequential information learning module, respectively. We adopt Adam as the optimizer with a learning rate of 0.001. We train for 200 epochs on each dataset. L2 loss weight is set 5e-4 for R8 and 20NG, 5e-5 for others. To evaluate the classification performance of all approaches, we adopt Accuracy (ACC) as the evaluation metric.

\subsection{Learning Performance Comparisons}
In this subsection, we compared our method with the state-of-the-art graph-based text classification approaches on five datasets, fully demonstrating its effectiveness in text classification tasks.
The average classification performance evaluated by ACC has been listed in Table \ref{syjg}.
Compared to graph-based approaches, our approach achieves significant improvements.
On short-to-medium text datasets R8, R52, and MR, our model achieves improvements of 0.02\%, 0.6\%, and 1.9\% over the strongest baseline, respectively. On long-text datasets, Ohsumed and 20NG, our model outperforms the best baseline with improvements of 1.8\% and 0.6\%, respectively. 
Compared to TextING, which relies on sliding-window text-level graphs, our model is capable of learning diverse structural relationships within the text-level graph. 
Therefore, the classification performance of our model outperforms TextING across all datasets.
In addition, our approach overcomes the limitation of conventional graph models that neglect long-range sequential information. 
The integration of long-range sequential information with diverse structural information allows the model to learn more discriminative text representations, thus outperforming most graph-based methods. 
We will further reveal the effects of co-occurrence, syntax, and semantic relationships on learning more discriminative text representations in Section \ref{relation}.

\begin{figure}[t]\renewcommand{\arraystretch}{1.5}
	\centering
\includegraphics[width=0.64\textwidth]{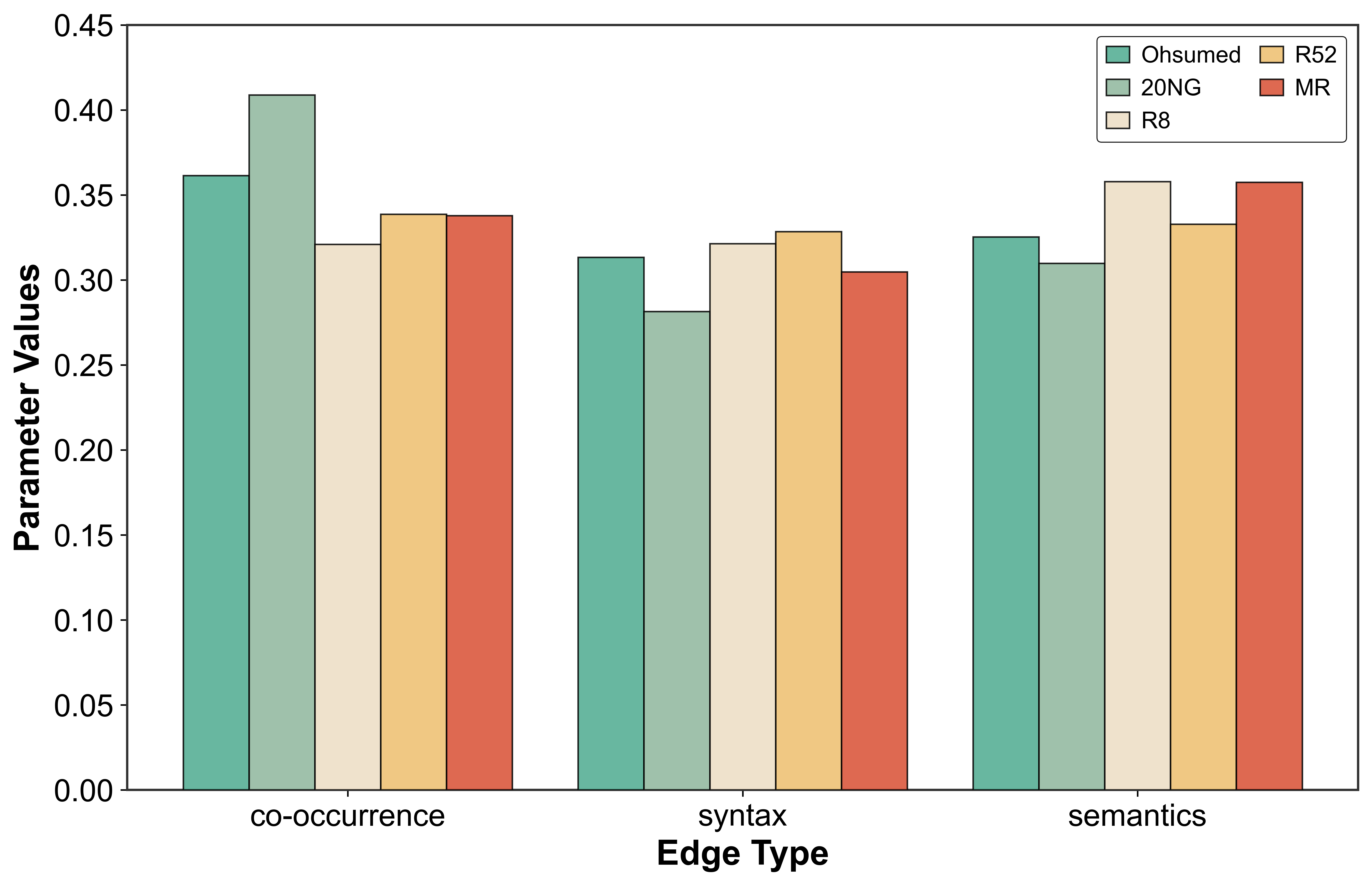}
	\caption{Visualization of adaptive parameters for co-occurrence, syntax, and semantic relations across five text datasets. Specifically, for long-text datasets, co-occurrence relations are more important for learning more discriminative text representations.}
	\label{para}
\end{figure}

\renewcommand{\arraystretch}{1.3}
\begin{table*}[t]
\caption{Performance comparisons (ACC) of different variants of TextGSL.
}
\label{as1}
\begin{center}
\normalsize \centering \begin{tabular}{p{2cm}|p{2.2cm} p{2.2cm} p{2.2cm} p{2.2cm} p{2.2cm}}
\hline
\textbf{Components}& \textbf{R8} & \textbf{R52} & \textbf{MR} & \textbf{Ohsumed} & \textbf{20NG}\\
\hline
\hline
w/o LSL & 0.9814 \small ± 0.0009 & 0.9607 \small ±  0.0007 & 0.8102 \small ±  0.0019 & 0.7132 \small ±  0.0021 & 0.8751 \small ±  0.0005\\

w/o DSL & 0.9782 \small ± 0.0015 & 0.9561 \small ± 0.0004 & 0.8073 \small ± 0.0014 & 0.7148 \small ± 0.0013 & 0.8765 \small ± 0.0011\\			
\hline
TextGSL & \textbf{0.9834} \small ± 0.0006 & \textbf{0.9626} \small ± 0.0015 & \textbf{0.8127} \small ± 0.0021& \textbf{0.7216} \small ± 0.0030 & \textbf{0.8832} \small ± 0.0006 \\	

\hline
\end{tabular}
\end{center}
\end{table*}

\subsection{Diverse Relationship Analysis}\label{relation}
In this subsection, we visualize the adaptive parameters $\quad \ell \in \{ \text{co}, \text{syn}, \text{sem}\}$ to further reveal which type of relationship in the diverse structural information learning module can learn more discriminative text representations.
Fig.\ref{para} shows the adaptive parameter values across different datasets, demonstrating that the proposed model can effectively learn diverse structural information between word pairs.
For long-text datasets, Ohsumed and 20NG, co-occurrence relations exhibit the largest weights, while syntactic and semantic relations have relatively smaller weights. This is because the syntax and semantic relationships in long texts are overly complex and may introduce some redundant edges.
For short-text datasets, the syntactic relationships between word pairs are relatively simple, so the parameter value of co-occurrence and semantic relationships is slightly higher than syntactic relationships. 
TextGSL can learn more discriminative text representations from richer co-occurrence and syntactic relationships.

In order to demonstrate the impact of long-range sequential information and diverse structural information on learning high-quality text embeddings, we conducted ablation studies in Section \ref{ablation}.
\begin{figure}[ht] 
    \centering

    \begin{minipage}[b]{0.4\linewidth}
        \centering
        \includegraphics[width=\linewidth]{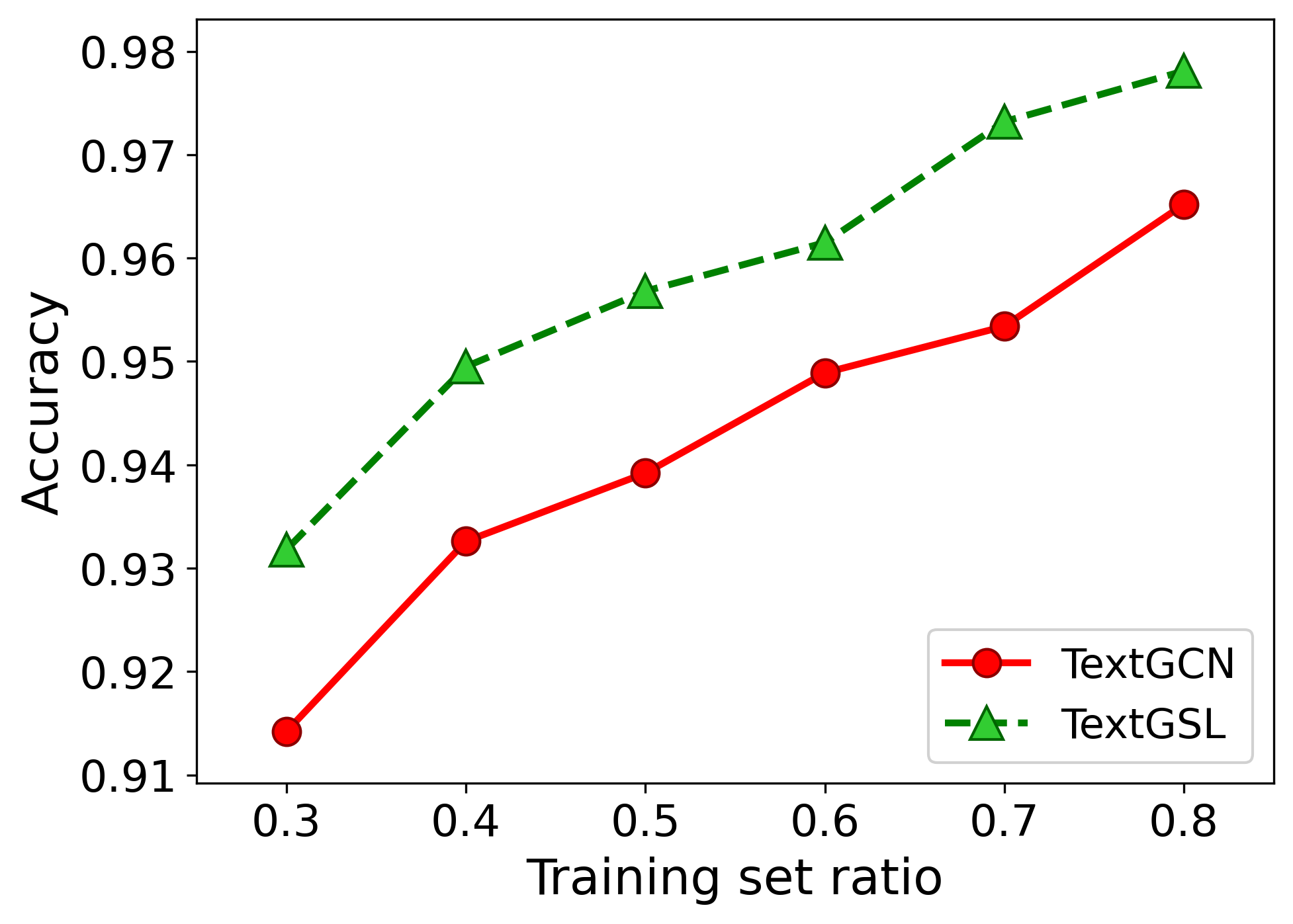}
        \\[2pt] 
        \footnotesize (a) R8 
    \end{minipage}\hspace{20pt}
    \begin{minipage}[b]{0.4\linewidth}
        \centering
        \includegraphics[width=\linewidth]{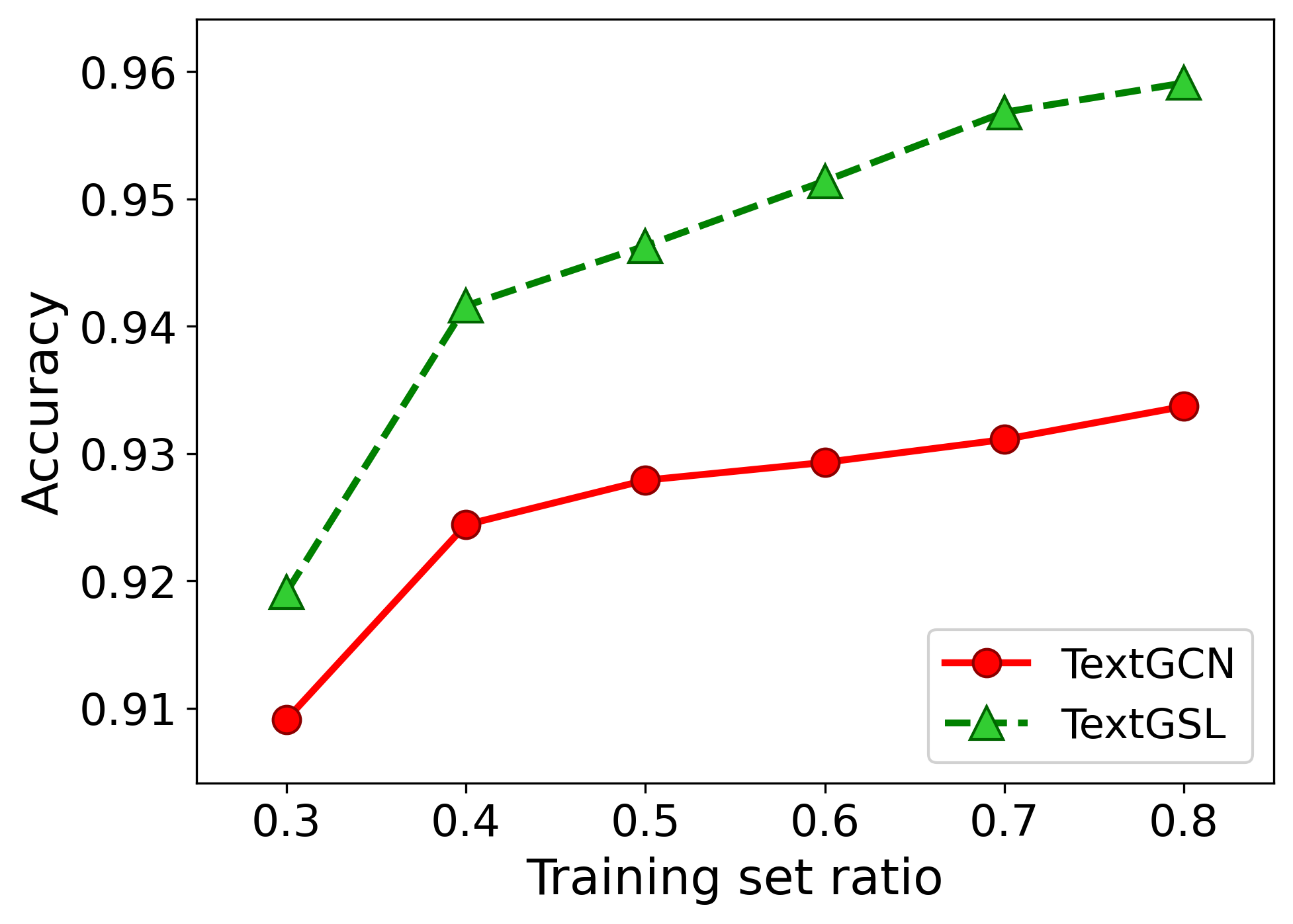}
        \\[2pt]
        \footnotesize (b) R52
    \end{minipage}

    \vspace{5pt}
    \begin{minipage}[b]{0.4\linewidth}
        \centering
        \includegraphics[width=\linewidth]{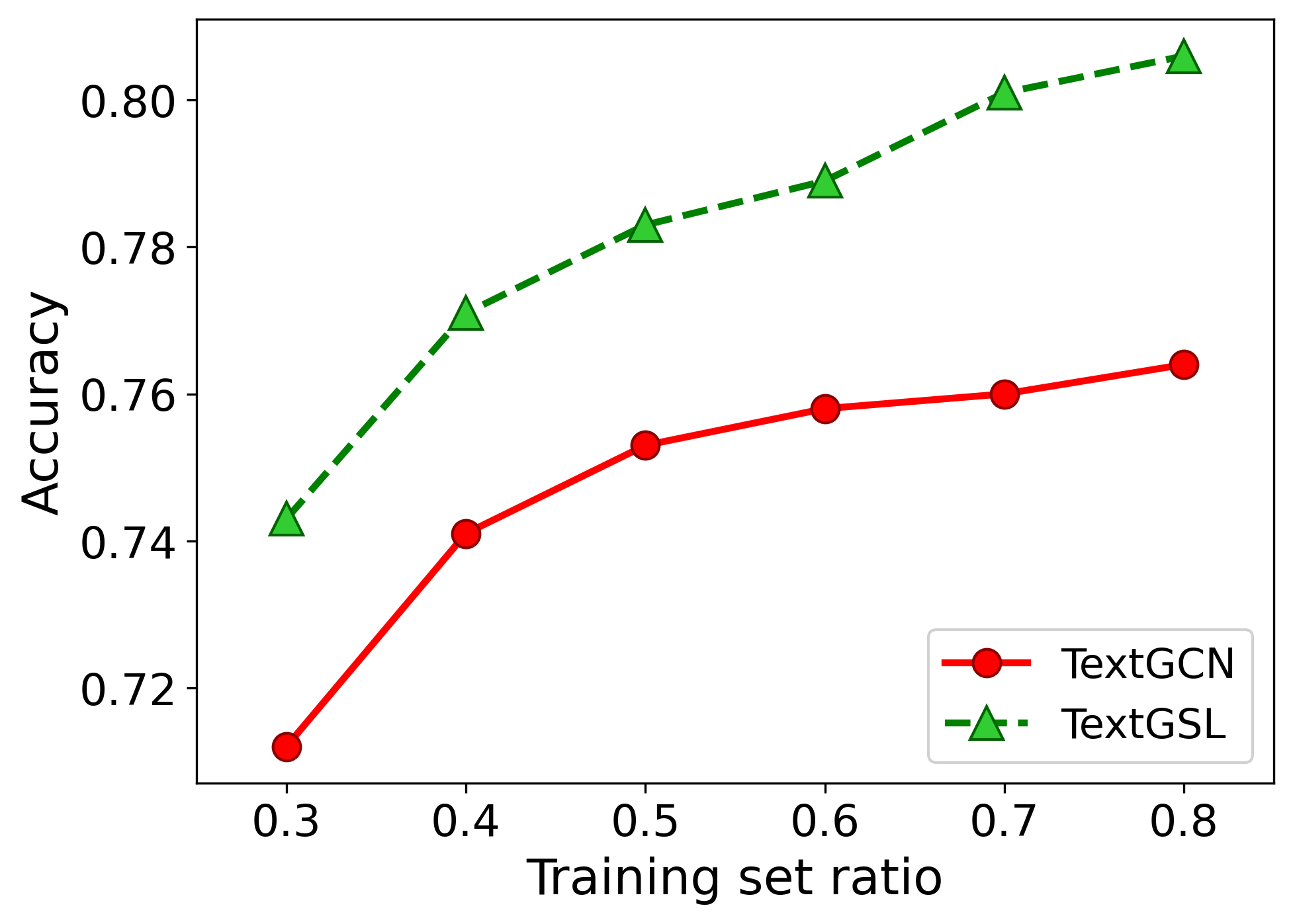}
        \\[2pt]
        \footnotesize (c) MR
    \end{minipage}\hspace{20pt}
    \begin{minipage}[b]{0.4\linewidth}
        \centering
        \includegraphics[width=\linewidth]{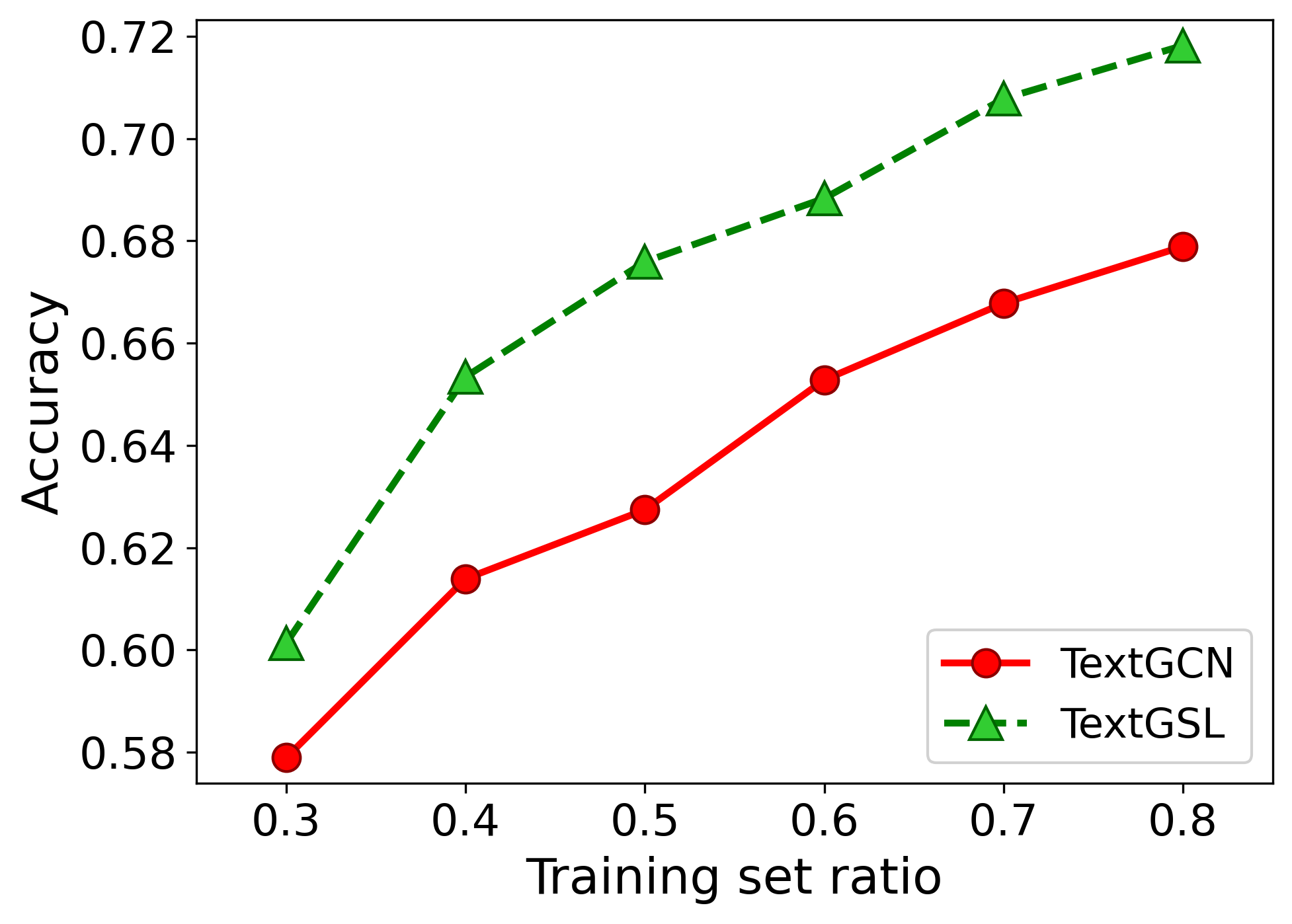}
        \\[2pt]
        \footnotesize (d) Ohsumed
    \end{minipage}

    \vspace{5pt}

    \begin{minipage}[b]{0.4\linewidth}
        \centering
        \includegraphics[width=\linewidth]{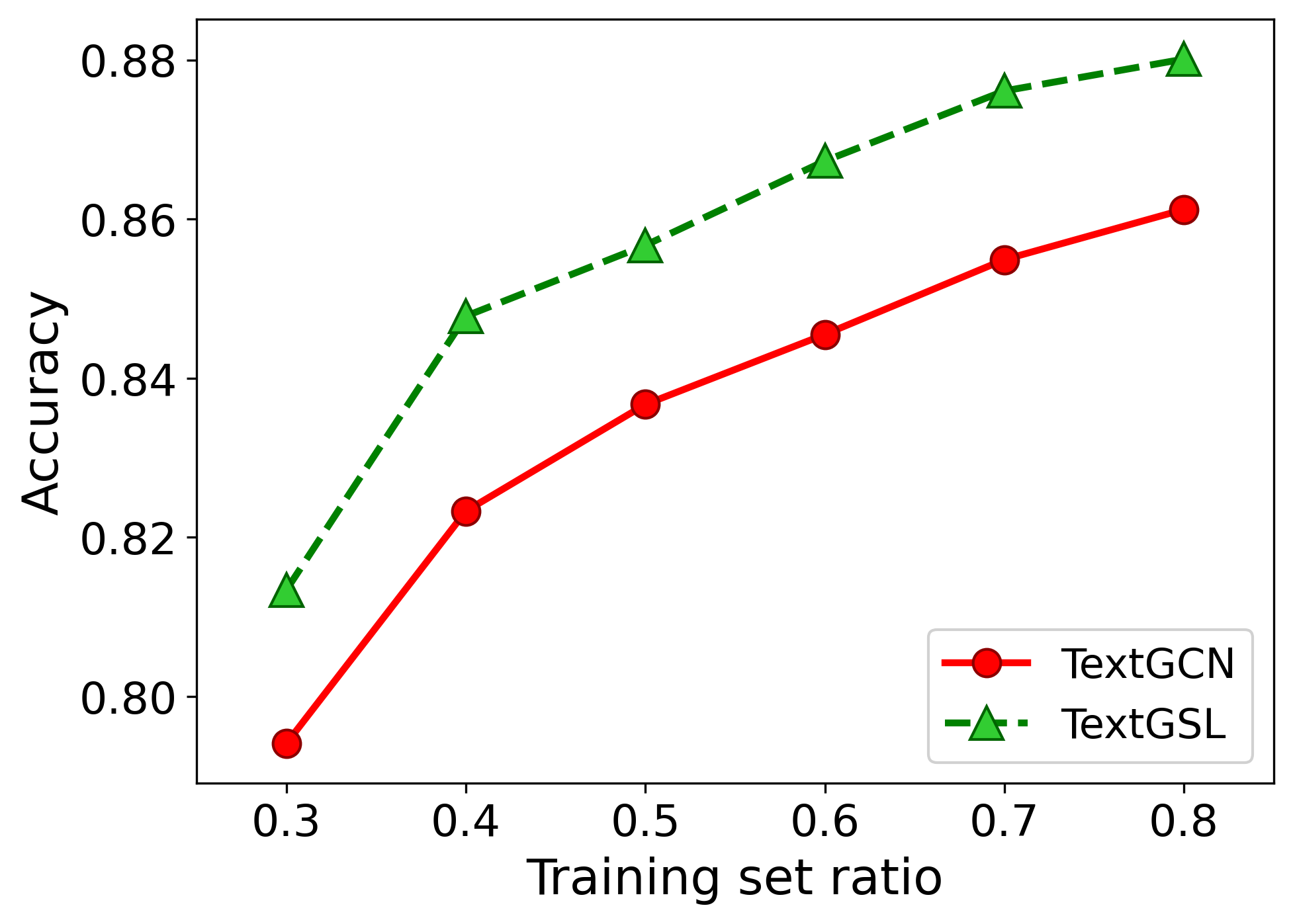}
        \\[2pt]
        \footnotesize (e) 20NG
    \end{minipage}
    
    \caption{Test accuracy by varying training set ratio}
    \label{all_datasets}
\end{figure}

\subsection{Ablation Studies}\label{ablation}
In this section, we conducted ablation studies to investigate the contribution of each module to the TextGSL model. 
We separately removed the long-range sequential information learning module (w/o LSL) and the diverse structural information learning module (w/o DSL). Then we conducted experiments to observe the model's classification performance. The average classification performance of ablation studies has been listed in Table \ref{as1}. 
Experiments show that removing the long-range sequence information learning module and the diverse structural information learning module leads to a decrease in accuracy.
Due to the over-smoothing limitation of GNNs, they are difficult to learn long-range dependencies.
However, transformers can capture information from long-range word nodes through a self-attention mechanism. 
Therefore, for long-text datasets like 20NG and Ohsumed, long-range sequential information can better learn more discriminative text representations. 
TextGSL outperforms w/o LSL by 1.2\% on the Ohsumed dataset and by 0.9\% on the 20NG dataset.

From the ablation results, we observe that local sequential information and diverse structural information bring non-trivial performance gain on most of the text datasets.

\subsection{ Effects of Different Training Set Ratio}\label{train}
This section discusses the effect of variations in the training set ratio on the model's learning capacity. 
TextGCN is a transductive learning model that is difficult to classify new text. 
To demonstrate the stability of our model's learning ability and its effectiveness in inductive text classification, 
we vary the ratio of the training set and compare the performance of our proposed model against TextGCN across the five datasets. 
Fig. \ref{all_datasets} reports test accuracies with 0.3, 0.4, 0.5, 0.6, 0.7, 0.8 training set ratios of five datasets. In the experiments, the graph structures from each text in the test set are unseen during training.
As the size of the training set increases, the accuracy improves consistently across almost all datasets.
Experiments show that, with a limited training set, our model outperforms the transductive learning model TextGCN. Our model can classify word nodes and text graph structures that it has never seen before.

\section{Conclusions}
In this paper, we propose TextGSL, which is a graph-sequence learning method for inductive text classification.
More specifically, we construct a text-level graph based on the different relationships (e.g., co-occurrence, syntax, semantics) between word pairs.
Then, we designed an adaptive multi-edge message-passing paradigm to learn the diverse structural information between word pairs. 
In order to overcome the over-smoothing problem of GNNs, we use a transformer layer to capture long-range sequential information.
The integration of long-range sequential information and diverse structural information enables models to learn more discriminative text representations.
The experimental results show that our proposed TextGSL achieves significant improvement in text classification tasks.
In the future, we will combine pre-trained models to further enhance the model's ability to learn high-quality text representations.

\bibliographystyle{IEEEtran}
\bibliography{ref}

\end{document}